\def\cvprPaperID{7821}
\def\confName{CVPR}
\def\confYear{2023}

\newif\ifreview 
\newif\ifarxiv 
\newif\ifcamera \newcommand{\cameraready}{\cameratrue}
\newif\ifrebuttal 

\cameraready
\pdfoutput=1
\documentclass[10pt,twocolumn,letterpaper]{article}
\ifreview \usepackage[review]{cvpr} \fi
\ifarxiv \usepackage[pagenumbers]{cvpr} \fi
\ifrebuttal \usepackage[rebuttal]{cvpr} \fi
\ifcamera \usepackage{cvpr} \fi

\usepackage{graphicx}
\usepackage{amsmath}
\usepackage{amssymb}
\usepackage{booktabs}
\usepackage{pifont}
\usepackage{bm}
\usepackage{algorithm}
\usepackage{algorithmic}

\usepackage{times}
\usepackage{microtype}
\usepackage{epsfig}
\usepackage[table,xcdraw]{xcolor}
\usepackage{caption}
\usepackage{float}
\usepackage{placeins}
\usepackage{color, colortbl}
\usepackage{stfloats}
\usepackage{enumitem}
\usepackage{tabularx}
\usepackage{xstring}
\usepackage{multirow}
\usepackage{xspace}
\usepackage{url}
\usepackage{subcaption}
\usepackage{xcolor}
\usepackage[hang,flushmargin]{footmisc}

\ifcamera \usepackage[accsupp]{axessibility} \fi

% ...

\ifarxiv  \fi

\newcommand{\R}[1]{{%
    \textbf{%
        \ifstrequal{#1}{1}{\textcolor{red}{R#1}}{%
        \ifstrequal{#1}{2}{\textcolor{blue}{R#1}}{%
        \ifstrequal{#1}{3}{\textcolor{magenta}{R#1}}{%
        \ifstrequal{#1}{4}{\textcolor{teal}{R#1}}{%
                           \textcolor{cyan}{R#1}%
        }}}}%
    }%
}}

\usepackage{xr-hyper}

\makeatletter
\newcommand*{\addFileDependency}[1]{
  \typeout{(#1)}
  \@addtofilelist{#1}
  \IfFileExists{#1}{}{\typeout{No file #1.}}
}

\makeatother

\usepackage[pagebackref,breaklinks,colorlinks]{hyperref}
\usepackage[capitalize]{cleveref}
\crefname{section}{Sec.}{Secs.}
\crefname{table}{Table}{Tables}
\crefname{figure}{Fig.}{Figs.}

\begin{document}
\newcommand{\yonggang}[2]{\textcolor{blue}{#1}} 

\title{Hard Sample Matters a Lot in Zero-Shot Quantization}
\author{
    Huantong Li\textsuperscript{\rm 1~\rm 6}~\thanks{Equal contribution. Email: 202021046173@mail.scut.edu.cn} ~~ 
    Xiangmiao Wu\textsuperscript{\rm 1}~~
    Fanbing Lv\textsuperscript{\rm 2}~~
    Daihai Liao\textsuperscript{\rm 2} \\
    Thomas H. Li\textsuperscript{\rm 7}~~
    Yonggang Zhang\textsuperscript{\rm 3}~\footnotemark[1]~~\footnotemark[2]~~
    Bo Han\textsuperscript{\rm 3}~~
    Mingkui Tan\textsuperscript{\rm 1~\rm 4~\rm 5}~\thanks{Corresponding author. Email: mingkuitan@scut.edu.cn} \\
    \textsuperscript{\rm 1}\small{South China University of Technology,}
    \textsuperscript{\rm 2}\small{Changsha Hisense Intelligent System Research Institute Co., Ltd} \\
    \textsuperscript{\rm 3}\small{Hong Kong Baptist University,}
    \textsuperscript{\rm 4}\small{Key Laboratory of Big Data and Intelligent Robot, Ministry of Education} \\
    \textsuperscript{\rm 5}\small{PengCheng Laboratory,}
    \textsuperscript{\rm 6}\small{Information Technology R\&D Innovation Center of Peking University} \\
    \textsuperscript{\rm 7}\small{School of Electronic and Computer Engineering, Peking University Shenzhen Graduate School, Shenzhen, China} \\
}
\maketitle

\begin{abstract}
Zero-shot quantization (ZSQ) is promising for compressing and accelerating deep neural networks when the data for training full-precision models are inaccessible. In ZSQ, network quantization is performed using synthetic samples, thus, the performance of quantized models depends heavily on the quality of synthetic samples. Nonetheless, we find that the synthetic samples constructed in existing ZSQ methods can be easily fitted by models. Accordingly, quantized models obtained by these methods suffer from significant performance degradation on hard samples.
To address this issue, we propose \textbf{HA}rd sample \textbf{S}ynthesizing and \textbf{T}raining (HAST). Specifically, HAST pays more attention to hard samples when synthesizing samples and makes synthetic samples hard to fit when training quantized models.
HAST aligns features extracted by full-precision and quantized models to ensure the similarity between features extracted by these two models. Extensive experiments show that HAST significantly outperforms existing ZSQ methods, achieving performance comparable to models that are quantized with real data.


\end{abstract}
\section{Introduction}
\label{sec:intro}
\par Deep neural networks (DNNs) achieve great success in many domains, such as image classification~\cite{DBLP:journals/cacm/KrizhevskySH17,DBLP:journals/corr/SimonyanZ14a,DBLP:conf/cvpr/SzegedyLJSRAEVR15}, object detection~\cite{DBLP:conf/cvpr/GirshickDDM14,DBLP:conf/iccv/Girshick15,DBLP:conf/cvpr/PangCSFOL19}, semantic segmentation~\cite{DBLP:journals/ijcv/EveringhamEGWWZ15,DBLP:conf/cvpr/ZhuangSTL019}, and embodied AI~\cite{chen2022active,chen2022weakly,DBLP:journals/tetci/DuanYTZT22}. 
These achievements are typically paired with the rapid growth of parameters and computational complexity, making it challenging to deploy DNNs on resource-constrained edge devices. In response to the challenge, \emph{network quantization} proposes to represent the full-precision models, i.e., floating-point parameters and activations, using low-bit integers, resulting in a high compression rate and an inference-acceleration rate~\cite{DBLP:conf/cvpr/JacobKCZTHAK18}. These methods implicitly assume that the training data of full-precision models are available for the process of network quantization. However, the training data, e.g., medical records, can be inaccessible due to privacy and security issues. Such a practical and challenging scenario has led to the development of \emph{zero-shot quantization} (ZSQ)~\cite{ZeroQ}, quantizing networks without accessing the training data.


\begin{figure}[t]
\centering
\subfloat[3-bit quantization.]{\vspace{-2mm}\includegraphics[width=1.65in]{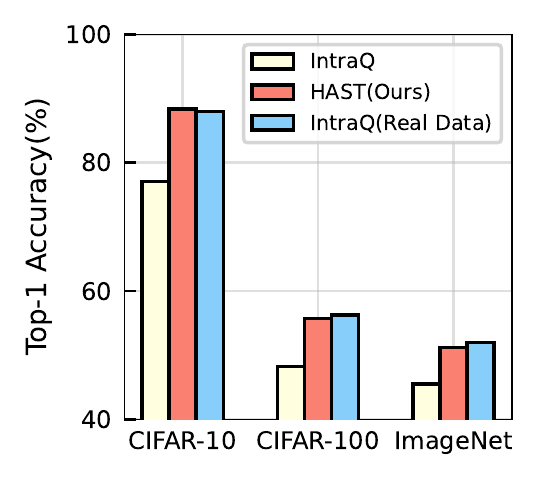}}
\subfloat[4-bit quantization.]{\vspace{-2mm}\includegraphics[width=1.65in]{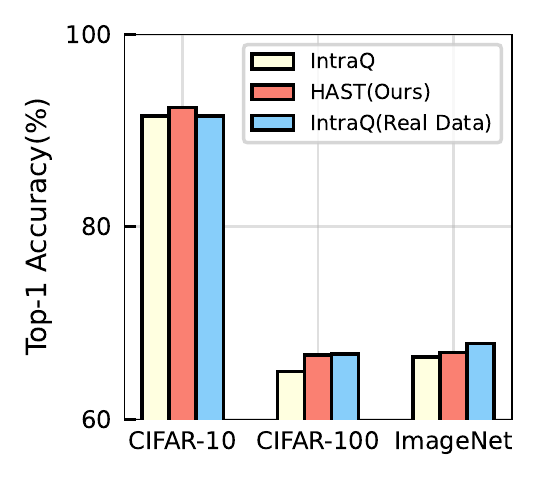}}
\vspace{-1mm}
\caption{Performance of the proposed HAST on three datasets compared with the state-of-the-art method IntraQ~\cite{IntraQ} and the method fine-tuning with real data~\cite{IntraQ}. HAST quantizes ResNet-20 on CIFAR-10/CIFAR-100 and ResNet-18 on ImageNet to 3-bit (left) and 4-bit (right), achieving performance comparable to the method fine-tuning with real data.}
\vspace{-5mm}
\label{Fig1}
\end{figure}

\par 
Many efforts have been devoted to ZSQ~\cite{DFQ,SQuant,ZeroQ,DSG,IntraQ,GDFQ}. In ZSQ, some works perform network quantization by weight equalization~\cite{DFQ}, bias correction~\cite{DBLP:conf/nips/BannerNS19}, or weight rounding strategy~\cite{SQuant}, at the cost of some performance degradation. To promote the performance of quantized models, advanced works propose to leverage synthetic data for network quantization~\cite{ZeroQ,DSG,IntraQ,GDFQ,AIT}. Specifically, they fine-tune quantized models with data synthesized using full-precision models, achieving promising improvement in performance.

Much attention has been paid to the generation of synthetic sample, since high-quality synthetic samples lead to high-performance quantized models~\cite{ZeroQ,GDFQ}. Recent works employ generative models to synthesize data with fruitful approaches, considering generator design~\cite{AutoReCon}, boundary sample generation~\cite{Qimera}, adversarial training scheme~\cite{ZAQ}, and effective training strategy~\cite{AIT}. Since the quality of synthetic samples is typically limited by the generator~\cite{Survey}, advanced works treat synthesizing samples as a problem of noise optimization~\cite{ZeroQ}. Namely, the noise distribution is optimized to approximate some specified properties of real data distributions, such as batch normalization statistics (BNS) and inception loss (IL)~\cite{inception}. To promote model performance, IntraQ~\cite{IntraQ} focuses on the property of synthetic samples and endows samples with heterogeneity, achieving state-of-the-art performance as depicted in Figure~\ref{Fig1}.

\begin{figure*}[t]
\centering
\subfloat[Performance of converged 3-bit ResNet-20.]{\vspace{-2mm}
\label{Fig2.a}
\includegraphics[width=2in]{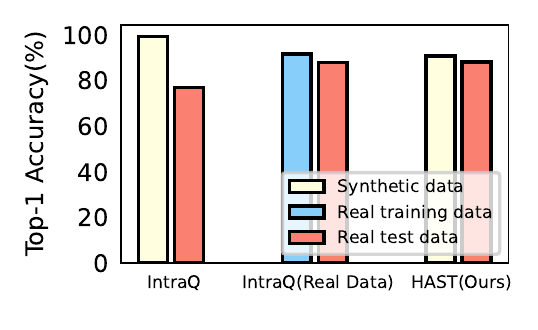}
}
\subfloat[Variation of error rate with sample difficulty.]{\vspace{-2mm}
\label{Fig2.b}
\includegraphics[width=1.96in]{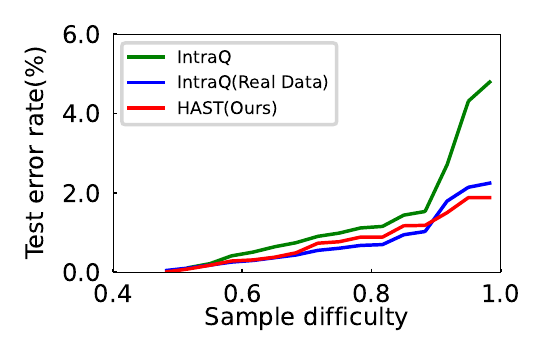}
}
\subfloat[Variation of fraction with sample difficulty.]{\vspace{-2mm}
\label{Fig2.c}
\includegraphics[width=2in]{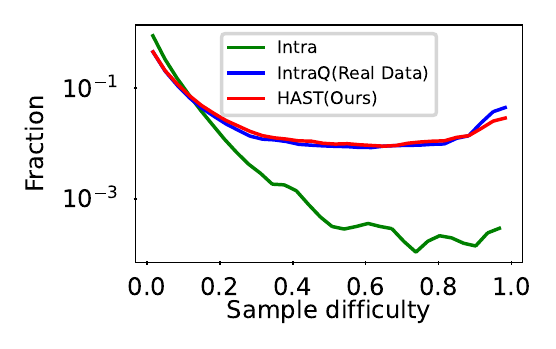}
}
\vspace{-1mm}
\caption{Analysis on synthetic data. (a) Performance of converged 3-bit ResNet-20. We quantize ResNet-20 to 3-bit using IntraQ~\cite{IntraQ}, Real Data~\cite{IntraQ}, and Our HAST, respectively, The top-1 accuracy on both training data (synthetic data for ZSQ methods) and test data is reported. (b) The error rate of test samples with different difficulties. (c) Distribution visualization of sample difficulty using GHM~\cite{GHM}. For each converged quantized model, we randomly sample 10,000 synthetic/real samples and count the fraction of samples based on difficulty. Note that the y-axis uses a log scale since the number of samples with different difficulties can differ by order of magnitude.}
\vspace{-3mm}
\label{Fig2}
\end{figure*}

Although existing ZSQ methods achieve considerable performance gains by leveraging synthetic samples, there is still a significant performance gap between models trained with synthetic data and those trained with real data~\cite{IntraQ}. To reduce the performance gap, we investigate the difference between real and synthetic data. Specifically, we study the difference in generalization error between models trained with real data and those trained with synthetic data. Our experimental results show that synthetic data lead to larger generalization errors than real data, as illustrated in Figure~\ref{Fig2.a}. Namely, synthetic sample lead to a more significant gap between training and test accuracy than real data. 

We conjecture that the performance gap stems from the misclassification of hard test samples. To verify the conjecture, we conduct experiments to study how model performance varies with sample difficulty, where GHM~\cite{GHM} is employed to measure the difficulty of samples quantitatively. The results shown in Figure~\ref{Fig2.b} demonstrate that, on difficult samples, models trained with synthetic data perform worse than those trained with real data.
This may result from that synthetic samples are easy to fit, which is consistent with the observation on inception loss of synthetic data~\cite{MixMix}. We verify the assumption through a series of experiments, where we count the fraction of samples of different difficulties using GHM~\cite{GHM}. The results are reported in Figure~\ref{Fig2.c}, where we observe a severe missing of hard samples in synthetic samples compared to real data. Consequently, quantized models fine-tuned with these synthetic data may fail to generalize well on hard samples in the test set. 

In light of conclusions drawn from Figure~\ref{Fig2}, the samples synthesized for fine-tuning quantized models in ZSQ should be hard to fit. To this end, we propose a novel \textbf{HA}rd sample \textbf{S}ynthesizing and \textbf{T}raining (HAST) scheme. The insight of HAST has two folds: a) The samples constructed for fine-tuning models should not be easy for models to fit; b) The features extracted by full-precision and the quantized model should be similar. To this end, in the process of synthesizing samples, HAST pays more attention to hard samples in a re-weighting manner, where the weights are equal to the sample difficulty introduced in GHM~\cite{GHM}. Meanwhile, in the fine-tuning process, HAST further promotes the sample difficulty on the fly and aligns the features between the full-precision and quantized models. 

To verify the effectiveness of HAST, we conduct comprehensive experiments on three datasets under two quantization precisions, following settings used in~\cite{IntraQ}. Our experimental results show that HAST using only 5,120 synthetic samples outperforms previous state-of-the-art method~\cite{IntraQ} and even achieves performance comparable with quantization with real data. 

Our main contributions can be summarized as follows:
\vspace{-1mm}
\begin{itemize}
    \item[$\bullet$] 
    We observe that the performance degradation in zero-shot quantization is attributed to the lack of hard samples. Namely, the synthetic samples used in existing ZSQ methods are easily fitted by quantized models, distinguishing models trained on synthetic samples from those trained on real data, as depicted in Figure~\ref{Fig2}.
    \vspace{-1mm}
    \item[$\bullet$] 
    Built upon our empirical observation, we propose a novel \textbf{HA}rd sample \textbf{S}ynthesizing and \textbf{T}raining (HAST) scheme to promote the performance of ZSQ. Specifically, HAST generates hard samples and further promotes the sample difficulty on the fly when training models, paired with a feature alignment constraint to ensure the similarity of features extracted by these two models, as summerized in Algorithm~\ref{alg:algorithm1}.
    \vspace{-1mm}
    \item[$\bullet$] 
    Extensive experiments demonstrate the superiority of HAST over existing ZSQ methods. More specifically, HAST using merely 5,120 synthetic samples outperforms the previous state-of-the-art method and achieves performance comparable to models fine-tuned using real training data, as shown in Figure~\ref{Fig1}.
\end{itemize}

\section{Related Work}
In this section, we briefly review the most relevant works in network quantization.

\label{sec:related}
\subsection{Data-Driven Quantization}
Network quantization is proposed to promote the compression rates and accelerate the inference by representing the full-precision models using low-bit integers. The straightforward and effective approach is to fine-tune models using the training data of full-precision models. In this regard, quantization aware training (QAT) methods focus on designing quantizers~\cite{PACT, QIL, LSQ, APoT, DSQ}, training strategies~\cite{DBLP:conf/cvpr/ZhuangLTSR20,DBLP:conf/cvpr/Lee0H21}, and binary networks~\cite{DBLP:conf/nips/LinJX00WHL20,DBLP:conf/iclr/MartinezYBT20,DBLP:conf/cvpr/QinGLSWYS20}.

In many practical scenarios, the training data of full-precision models can be inaccessible, limiting the effectiveness of QAT methods. To address the challenge, post-training quantization (PTQ) methods perform network quantization with limited training data~\cite{DBLP:conf/nips/BannerNS19,BRECQ}. Specifically, these methods approximate an optimal clipping value in the feature space using limited training samples~\cite{DBLP:conf/nips/BannerNS19} and introduce an allocation policy to quantize both activations and weights to 4-bit. However, both QAT and PTQ methods require training data to perform network quantization.


\subsection{Zero-Shot Quantization}
\par To further relax the privacy or security issue, ZSQ methods propose to perform network quantization without accessing the training data of full-precision models. A simple yet effective approach is calibrating model parameters without training data. For example, DFQ~\cite{DFQ} equalizes the weight ranges in the network and utilizes the scale and shift parameters stored in the batch normalization layers to correct biased quantization errors. SQuant~\cite{SQuant} performs data-free quantification using the diagonal Hessian approximation layer by layer. However, these methods may lead to significant performance degradation when quantizing models with ultra-low precision~\cite{IntraQ}.

Advanced works propose to generate synthetic samples for fine-tuning quantized models, resulting in promoted model performance. GDFQ~\cite{GDFQ} first adopts generative models guided by both the batch normalization statistics and extra category label information to synthesize samples. Since the performance depends heavily on synthetic samples, many variants of GDFQ improve the performance by adopting better generator~\cite{AutoReCon}, adversarial training~\cite{ZAQ}, boundary-supporting samples generation~\cite{Qimera}, and  using an ensemble of compressed models~\cite{GZNQ}. Recently, it has been shown that data synthesis can be realized by optimizing random noise sampled from a pre-defined distribution~\cite{ZeroQ}. The noise optimization scheme is further improved by carefully designing an appropriate mechanism for synthesizing data, such as enhancing the diversity of samples~\cite{DSG} and increasing the intra-class heterogeneity of synthetic samples~\cite{IntraQ}. Although these methods achieve considerable performance gain compared to data-driven quantization, there is still a performance gap between fine-tuning with synthetic and real data.


\section{Methodology}
\label{sec:method}
\par
\begin{figure}[t]
\centering
\includegraphics[width=\linewidth]{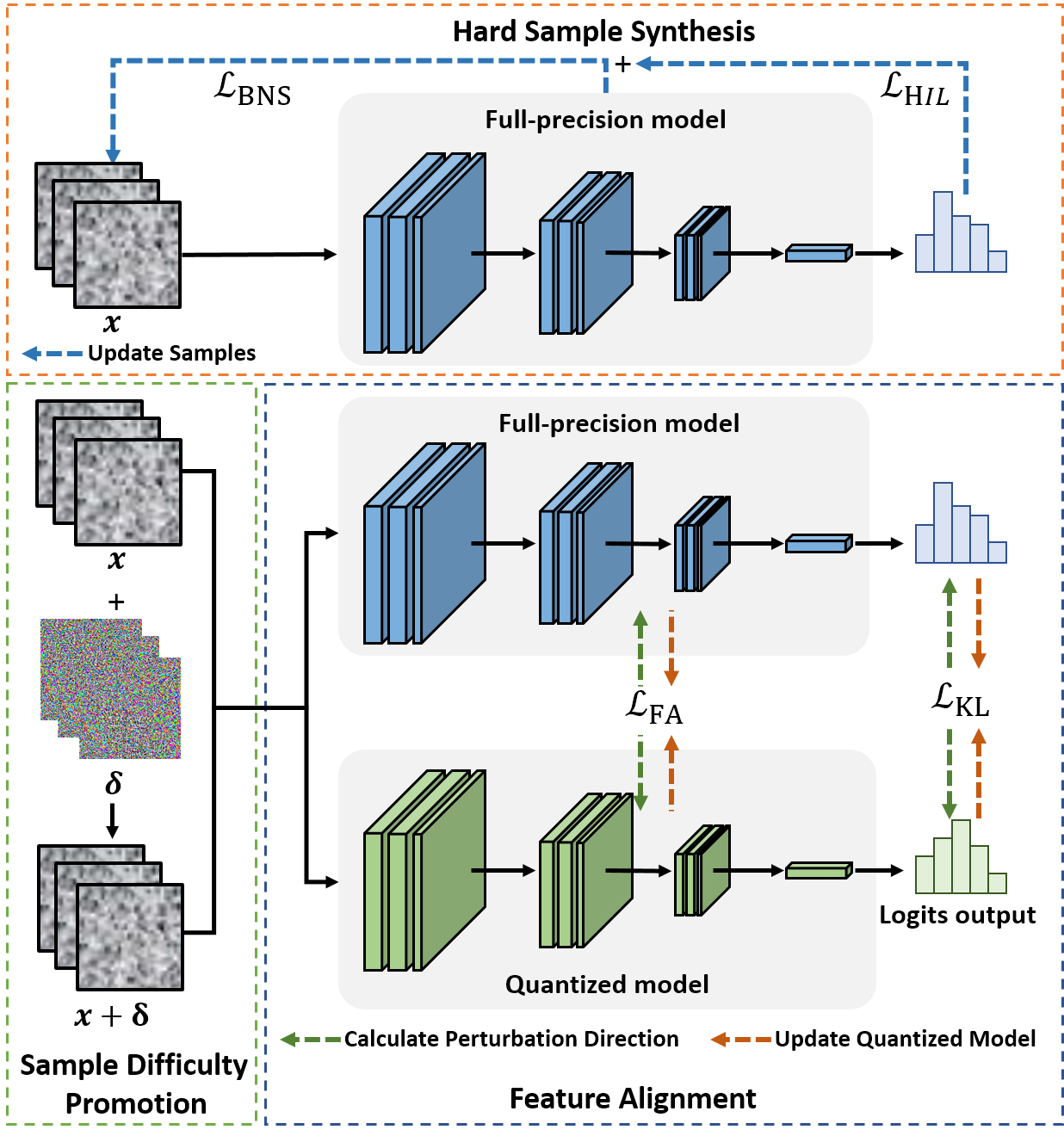}
\vspace{-3mm}
\caption{An overview of the proposed hard sample synthesizing and training (HAST) scheme. The hard sample synthesis focuses on generating hard samples in the data generation process. The sample difficulty promotion makes synthetic samples hard to fit. The feature alignment guides the quantized model to handle hard samples.}
\vspace{-4.9mm}
\label{Fig3}
\end{figure}

 In this section, we first introduce the process of zero-shot quantization (ZSQ). Then, we detail the proposed \textbf{HA}rd sample \textbf{S}ynthesizing and \textbf{T}raining (HAST) scheme containing three parts: \emph{hard sample synthesis}, \emph{sample difficulty promotion}, and \emph{feature alignment} ensuring the similarity of representations from quantized models and full precision models. The overview of HAST is illustrated in Figure~\ref{Fig3}.

\subsection{Preliminaries}
\label{Preliminaries}
\par \textbf{Quantizer.} We focus on the asymmetric uniform quantizer to implement network quantization in this work, following~\cite{GDFQ, IntraQ}. Denoting $\theta$ as weights of a full-precision model, $l$ and $u$ as the lower and upper bound of $\theta$, and $n$ as the bit-width, the quantizer produces the quantized integer $\theta^q$ as follows:
\begin{equation}
\begin{aligned}
\label{eq1} \theta^q=\lfloor\theta\times S -z\rceil, 
\ 
S=\frac{2^n-1}{u-l}, \ 
z=S \times l+ 2^{n-1}
\end{aligned}
\end{equation}
where $S$ is the scaling factor to convert the range of $\theta$ to $n$-bit, $z$ decides which quantized value zero is mapped to, and ``$\lfloor\rceil$'' is an operation used to round its input to the nearest integer. Accordingly, the corresponding dequantized value can be calculated as:
\begin{equation}
\begin{aligned}
\label{eq2} \theta^{\prime}=\frac{\theta^q+z}{S}.
\end{aligned}
\end{equation}
Built upon the quantizer, the weights of full-precision models can be represented using low-bit integers. Consequently, the quantized model performs inference using the dequantized parameter $\theta^{\prime}$, which is typically paired with considerable performance degradation.

\textbf{Zero-Shot Quantization.} To mitigate the performance degradation, a widely adopted approach is to optimize $\theta$ such that its dequantized parameter $\theta^{\prime}$ can perform well on test sets~\cite{PACT}. This is typically achieved by optimizing $\theta$ using the data for training full-precision models~\cite{LSQ}, since $\theta^{\prime}$ is inherently the function of $\theta$. Zero-Shot Quantization (ZSQ) takes a step further to quantize models when data for training full-precision models are inaccessible.

In ZSQ, synthetic samples are usually employed for optimizing quantized models~\cite{GDFQ,ZeroQ}. The synthetic samples can be derived by noise optimization~\cite{ZeroQ,DSG,IntraQ}, which is typically instantiated by distribution approximation~\cite{ZeroQ,GDFQ}. Given a set of noise $\{\mathbf{x}_{i} \}^{N}_{i=1}$, synthetic samples are obtained by optimizing these noise to match the batch normalization statistics (BNS):
\begin{equation}
\begin{aligned}
\label{eq3} 
\min_{\{\mathbf{x}_{i} \}^{N}_{i=1}} \mathcal{L}_{BNS}
\triangleq
\frac{1}{L}
\sum_{l=1}^{L}
(
&||\mu^{l}(\theta) - 
\mu^{l}(\theta, \{\mathbf{x}_{i} \}^{N}_{i=1})|| \\
+
&||\sigma^{l}(\theta) - 
\sigma^{l}(\theta, \{\mathbf{x}_{i} \}^{N}_{i=1})||
),
\end{aligned}
\end{equation}
where $\mu^{l}(\theta)/\sigma^{l}(\theta)$ are mean/variance parameters stored in the $l$-th BN layer of full-precision model parameterized with $\theta$ and $\mu^{l}(\theta, \{\mathbf{x}_{i} \}^{N}_{i=1})/\sigma^{l}(\theta, \{\mathbf{x}_{i} \}^{N}_{i=1})$ are mean/variance parameters calculated on the sampled noise using $\theta$. Besides the BNS alignment objective function, an inception loss is also employed for optimizing sampled noise:
\begin{equation}
\begin{aligned}
\label{eq4} 
\min_{\{\mathbf{x}_{i} \}^{N}_{i=1}} \mathcal{L}_{IL}
\triangleq
\frac{1}{N}
\sum_{i=1}^{N}
CE(p(\mathbf{x}_i;\theta), \mathbf{y}_i)
,
\end{aligned}
\end{equation}
where $p(\cdot;\theta)$ stands for the probability predicted by full-precision model parameterized with $\theta$, $CE(\cdot,\cdot)$ represents the cross-entropy loss, and $\mathbf{y}_i$ is the label assigned to $\mathbf{x}_i$ as a prior classification knowledge. Consequently, synthetic data are obtained by optimizing the final objective function composed of these two terms:
\begin{equation}
\begin{aligned}
\label{eq5} 
\min_{\{\mathbf{x}_{i} \}^{N}_{i=1}} \mathcal{L}_{FNL}
\triangleq
\mathcal{L}_{BNS} + \beta\mathcal{L}_{IL}
,
\end{aligned}
\end{equation}
where $\beta$ is a hyper-parameter balancing the importance of two terms. Built upon the objective function $\mathcal{L}_{FNL}$, ZSQ can fine-tune the model parameter $\theta$ such that its dequantized parameter $\theta^{\prime}$ can perform well on synthetic samples. More specifically, the quantized network is usually fine-tuned by a teacher-student framework with the cross-entropy loss $CE$ and the Kullback-Leibler loss $KL$:
\begin{equation}
\begin{aligned}
\label{eq6} 
\min_{\theta^{\prime}}
\frac{1}{N}
\sum_{i=1}^{N}
CE(p(\mathbf{x}_i;\theta^{\prime}),
\mathbf{y}_i)
+
\alpha
KL(
p(\mathbf{x}_i;\theta)||
p(\mathbf{x}_i;\theta^{\prime})
),
\end{aligned}
\end{equation}
where $\alpha$ is a hyper-parameter balancing the importance of two terms. We use $\theta^{\prime}$ to represent the parameter of quantized models and optimize $\theta^{\prime}$. This is equal to optimizing $\theta$, as $\theta^{\prime}$ is a function of $\theta$ according to Eq.~(\ref{eq2}).

\subsection{General Scheme of Proposed Methods}

The detailed procedure of the proposed HAST scheme is summarized in Algorithm~\ref{alg:algorithm1}. HAST consists of three parts: hard sample synthesis, sample difficulty promotion, and feature alignment that is illustrated in Sec.~\ref{sec:method}.

In the process of sample synthesis, we start with a batch of random input $\mathbf{x}_i$ sampled from a standard Gaussian distribution, following~\cite{IntraQ}. The proposed hard sample synthesis, i.e., Eq. (\ref{HFNL}), optimizes $\mathbf{x}_i$, aiming to increase the sample difficulty measured with full-precision models. As shown in the top of Figure \ref{Fig3}, we feed $\mathbf{x}_i$ into the full-precision model and optimize it using the proposed hard-sample-enhanced final loss for synthesizing hard samples, i.e., Eq.~(\ref{HFNL}) by computing the BNS alignment loss (Eq.~(\ref{eq3})) and the hard-sample-enhanced inception loss (Eq.~(\ref{hil})). 

In the process of network fine-tuning, we increase the sample difficulty on the fly using Eq.~(\ref{SDP}), as shown in the bottom left of Figure~\ref{Fig3}. Then, we perform feature alignment to ensure the similarity between the full-precision and quantized models, as shown in the bottom right of Figure~\ref{Fig3}.

\begin{algorithm}[t]
	\caption{Hard Sample Synthesizing and Training for Zero-shot Quantization.}
	\label{alg:algorithm1}
	\textbf{Input:}{Pretrained full-precision model with parameter $\theta$; Initial synthetic dataset $D_s=\varnothing$; Number of synthetic images $N$; Generation iterations $T_{generate}$; Fine-tuning iterations $T_{fine-tune}$.}\\
	\textbf{Output:}{Low precision quantized model with parameter $\theta^{\prime}$.}
	\begin{algorithmic}
	\WHILE{$|D_s|\leq{N}$}
	\STATE{Sample batch of Gaussian noise $\mathbf{x}_i$ and label $y$;}
	\FOR{$t=1,...,T_{generate}$}
	\STATE{Optimize $\mathbf{x}_i$ by minimizing Eq.~(\ref{HFNL});}
	\ENDFOR
	\STATE $D_s \gets D_s\cup{\mathbf{x}_i}$
	\ENDWHILE
	\STATE{Get Synthetic dataset $D_s$.}\\
	\FOR{$t=1,...,T_{fine-tune}$}
	\STATE{Sample batch $\mathbf{x}_i$ from synthetic dataset $X$;}
	\STATE{Compute the perturbation $\delta$ by Eq.~(\ref{SDP});}
	\STATE{Update quantized model by minimizing Eq.~(\ref{eq:FAR});}
	\ENDFOR
	\STATE{Get Converged quantized model.}
	\end{algorithmic}
\end{algorithm}

\subsection{Hard Sample Synthesis}
\label{HSS}
\par Inspired by our experimental results in Figure~\ref{Fig2}, we propose to reconsider the difficulty of synthetic samples. Specifically, we increase the importance of hard samples while suppressing the importance of easy-to-fit samples, employing the GHM introduced in~\cite{GHM} to measure the sample difficulty $d$ of $\mathbf{x}$:
\begin{equation} \label{eq111}
d(\mathbf{x}, \theta) = 1 - p_{\mathbf{y}}(\mathbf{x}, \theta)
\end{equation}
where $p_{\mathbf{y}}(\mathbf{x})$ is the probability on label $\mathbf{y}$ predicted by the model parameterized with $\theta$. Through Eq.~\eqref{eq111}, we hope models focus more on transferable components~\cite{What_Transferred_Dong_CVPR2020}.

\begin{figure}[t]
\centering
\includegraphics[width=0.8\linewidth]{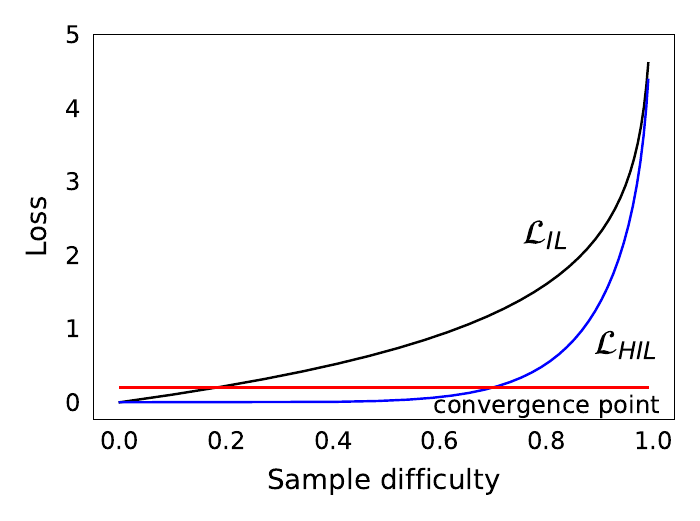}
\vspace{-4mm}
\caption{Curves of $\mathcal{L}_{IL}$ and $\mathcal{L}_{HIL}$.}
\vspace{-7mm}
\label{Fig4}
\end{figure}

Built upon the sample difficulty $d(\mathbf{x}, \theta)$, we propose a hard-sample-enhanced inception loss $\mathcal{L}_{HIL}$ to synthesize samples:
\begin{equation}
\begin{aligned}
\label{hil} 
\min_{\{\mathbf{x}_{i} \}^{N}_{i=1}} \mathcal{L}_{HIL}
\triangleq
\frac{1}{N}
\sum_{i=1}^{N}
d(\mathbf{x}_i, \theta)^\gamma
CE(p(\mathbf{x}_i;\theta), \mathbf{y}_i)
\end{aligned}
\end{equation}
where the hyper-parameter $\gamma$ controls how strongly the importance of hard samples is enhanced. The hard-sample-enhanced inception loss $\mathcal{L}_{HIL}$, i.e., Eq.~(\ref{hil}), pays more attention to hard samples in the synthesis process than the original inception loss $\mathcal{L}_{IL}$. Figure~\ref{Fig4} shows the curves of $\mathcal{L}_{IL}$ and $\mathcal{L}_{HIL}$. As sample difficulty decreases, loss drops and finally converges to zero. However, the sample optimized using $\mathcal{L}_{HIL}$ (0.7) is more difficult than that using $\mathcal{L}_{IL}$ (0.2). Consequently, the fraction of hard samples will increase, as depicted in Figure~\ref{Fig2.c}, where samples are synthesized by optimizing the hard-sample-enhanced loss:
\begin{equation}
\begin{aligned}
\label{HFNL} 
\min_{\{\mathbf{x}_{i} \}^{N}_{i=1}} \mathcal{L}_{HFNL}
\triangleq
\mathcal{L}_{BNS} + \beta\mathcal{L}_{HIL}
.
\end{aligned}
\end{equation}

\subsection{Sample Difficulty Promotion}
\label{sec:SDP}
Samples synthesized using Eq.~(\ref{HFNL}) are hard to fit for the full-precision model parameterized with $\theta$, but their difficulties for the quantized model $\theta^{\prime}$ are lacking. This stems from the fact that the difficulty used in Eq.~(\ref{HFNL}) is measured with $\theta$ using $d(\mathbf{x}_i, \theta)$. Therefore, we propose to promote the sample difficulty using the quantized model parameterized with $\theta^{\prime}$.

The sample difficulty promotion is different from the process of hard sample synthesis. To be specific, the latter utilizes full-precision models to synthesize samples for training quantized models, while the former leverages quantized models to enhance the sample difficulty for training themself. If the synthetic samples are changed in the process of sample difficulty promotion, the synthesized samples would be both easy and difficult for quantized models. Thus, we draw inspiration from adversarial training~\cite{FGSM,zhang2021causaladv}, where samples are modified to be hard to fit on the fly.

More specifically, we increase the sample difficulty $d(\mathbf{x}_i, \theta^{\prime})$ on the fly to make synthetic samples hard to fit by introducing a perturbation $\mathbf{\delta}_i$ for $\mathbf{x}_i$ in each training iteration:
\begin{equation}
\begin{aligned}
\label{SDP}
\mathbf{\delta}_i
=
\arg \max_{||\delta^{\prime}||_{\infty} \leq \epsilon}
d(\mathbf{x}_i + \delta_i^{\prime}, \theta^{\prime}),
\end{aligned}
\end{equation}
where $||\cdot||_{\infty}$ represents the $\ell_{\infty}$-norm and $\epsilon$ controls the strength of perturbation. We attach the perturbation $\delta_i$ for $\mathbf{x}_i$ to find nearby difficult samples with a larger sample difficulty in each iteration. 

\subsection{Minimizing Quantization Gap by Feature Alignment}
\label{FA}
Built upon the proposed hard sample synthesis and sample difficulty promotion, we are ready to train quantized models using hard samples. To ensure the quantized model performs the same as its full-precision model, an ideal result may be that the outputs of these two models (including intermediate features) are kept exactly the same, which shares the same spirit with~\cite{zhang2021causaladv} built upon a causal perspective. Along with this insight, we can align the features of the full-precision and quantized models:
\begin{equation}
\begin{aligned}
\label{eq:FA}
\min_{\theta^{\prime}}
\mathcal{L}_{FA}
\triangleq
\frac{\lambda}{NL}
\sum^{N}_{i=1} \sum^{L}_{l=1}
\phi^l(
f^l(\mathbf{x}_i + \delta_i;\theta), 
f^l(\mathbf{x}_i + \delta_i;\theta^{\prime})
),
\end{aligned}
\end{equation}
where $\phi^l(\cdot, \cdot)$ is a metric for the $l$-th layer measuring the difference of its inputs, e.g., mean square error, $f^l(\cdot;\theta)$ represents the feature at the $l$-th layer of a model parameterized with $\theta$, and the hyper-parameter $\lambda$ adjusts the order of magnitude of the loss.

The quantized models may not be able to match the outputs of full-precision models due to the limited model capacity. Thus, we relax the feature alignment constraint:
\begin{equation}
\begin{aligned}
\label{eq:FAR}
\min_{\theta^{\prime}}
\hat{\mathcal{L}}_{FA}
=
\frac{\lambda}{z}
\sum_{i , l \in S
}
& (
\phi(
f^l(\mathbf{x}_i + \delta_i;\theta), 
f^l(\mathbf{x}_i + \delta_i;\theta^{\prime})
) \\
+ & \alpha KL(p(\mathbf{x}_i; \theta) || p(\mathbf{x}_i; \theta^{\prime}))
)
,
\end{aligned}
\end{equation}
where $z=N |S|$, $S$ is the set of selected intermediate layers, $\phi(\cdot,\cdot)$ is the metric used for all intermediate layers, and $KL$ divergence is employed as the metric for the final output layer, i.e., the predicted probability. For the metric used in intermediate layers, we instantiate them with an attention vector~\cite{AttentionTransfer} rather than a mean square error ($|| f^l(\mathbf{x}_i + \delta_i;\theta) - f^l(\mathbf{x}_i + \delta_i;\theta^{\prime})||^2$). The metric of intermediate feature $f^l(\cdot;\theta) \in \mathbb{R}^{C \times W \times H}$ ($C, W$, and $H$ is the number of channels, width, and height) is calculated using an attention vector $Att(\cdot)$ as follows:
\begin{equation}
\begin{aligned}
\label{eq:att1} 
\phi(
f^l(\cdot;\theta), 
f^l(\cdot;\theta^{\prime})
) 
=
||Att(f^l(\cdot;\theta)) - Att(f^l(\cdot;\theta^{\prime}))||^2_2,
\end{aligned}
\end{equation}
where $Att(f^l(\cdot;\theta)) \in \mathbb{R}^{C}$ is a vector with the $c$-th element calculated as:
\begin{equation}
\label{eq:att2} 
Att(f^l)(c) = \sum_{w=1}^{W} \sum_{h=1}^{H} f^l(c,w,h)^2,
\end{equation}
where we denote $f^l(\cdot;\theta)$ by $f^l$ for simplicity.

\section{Experiments}
\label{sec:experiments}
In this section, we conduct extensive experiments to demonstrate the effectiveness of our proposed HAST scheme. We organize this section as follows. We first provide the experiment setting in Sec.~\ref{Experimental Setup}. Second, we compare our proposed method with existing ZSQ methods in Sec.~\ref{sec:cifar} and Sec.~\ref{sec:imagenet}. Last, we investigate the effect of the hyperparameters and the proposed component in our HAST scheme in Sec.~\ref{sec:ablation}. The code to reproduce our results is available \footnote{\url{https://github.com/lihuantong/HAST}}.

\begin{table}[t]   
\begin{center}   
\resizebox{0.48\textwidth}{!}{
\begin{tabular}{c|c|c|c}   
\hline 
Bit-width & Method & Generator & Acc.(\%) \\ \hline \hline
 & full-precision & - & 94.03 \\ \hline
\multirow{8}{*}{W4A4}& Real Data & - & 91.52 \\ \cline{2-4} 
\multirow{8}{*}{}& SQuant~\cite{SQuant} & - & 92.24 \\
\multirow{8}{*}{}& GDFQ~\cite{GDFQ} & \ding{52} & 90.25 \\
\multirow{8}{*}{}& AIT~\cite{AIT} & \ding{52} & 91.23 \\
\multirow{8}{*}{}& ZeroQ~\cite{ZeroQ}+IL~\cite{inception} & \ding{56} & 89.66 \\ 
\multirow{8}{*}{}& DSG~\cite{DSG}+IL~\cite{inception} & \ding{56} & 88.93 \\
\multirow{8}{*}{}& IntraQ~\cite{IntraQ} & \ding{56} & 91.49 \\
\multirow{8}{*}{}& HAST(Ours) & \ding{56} & \textbf{92.36}$\pm0.09$ \\ \hline

\multirow{8}{*}{W3A3}& Real Data & - & 87.94 \\ \cline{2-4} 
\multirow{8}{*}{}& SQuant~\cite{SQuant} & - & 79.19 \\
\multirow{8}{*}{}& GDFQ~\cite{GDFQ} & \ding{52} & 71.10 \\
\multirow{8}{*}{}& AIT~\cite{AIT} & \ding{52} & 80.49 \\
\multirow{8}{*}{}& ZeroQ~\cite{ZeroQ}+IL~\cite{inception} & \ding{56} & 69.53 \\ 
\multirow{8}{*}{}& DSG~\cite{DSG}+IL~\cite{inception} & \ding{56} & 48.99 \\
\multirow{8}{*}{}& IntraQ~\cite{IntraQ} & \ding{56} & 77.07 \\ 
\multirow{8}{*}{}& HAST(Ours) & \ding{56} & \textbf{88.34}$\pm0.06$ \\ \hline
\multicolumn{4}{c}{(a) Results on CIFAR-10.} \\

\multicolumn{4}{c}{  } \\ \hline

Bit-width & Method & Generator & Acc.(\%) \\ \hline \hline
 & full-precision & - & 70.33 \\ \hline
\multirow{8}{*}{W4A4}& Real Data & - & 66.80 \\ \cline{2-4} 
\multirow{8}{*}{}& SQuant~\cite{SQuant} & - & 63.96 \\
\multirow{8}{*}{}& GDFQ~\cite{GDFQ} & \ding{52} & 63.58 \\
\multirow{8}{*}{}& AIT~\cite{AIT} & \ding{52} & 65.80 \\
\multirow{8}{*}{}& ZeroQ~\cite{ZeroQ}+IL~\cite{inception} & \ding{56} & 63.97 \\ 
\multirow{8}{*}{}& DSG~\cite{DSG}+IL~\cite{inception} & \ding{56} & 62.62 \\
\multirow{8}{*}{}& IntraQ~\cite{IntraQ} & \ding{56} & 64.98 \\
\multirow{8}{*}{}& HAST(Ours) & \ding{56} & \textbf{66.68}$\pm0.12$ \\ \hline

\multirow{8}{*}{W3A3}& Real Data & - & 56.26 \\ \cline{2-4} 
\multirow{8}{*}{}& SQuant~\cite{SQuant} & - & 40.36 \\
\multirow{8}{*}{}& GDFQ~\cite{GDFQ} & \ding{52} & 43.87 \\
\multirow{8}{*}{}& AIT~\cite{AIT} & \ding{52} & 48.64 \\
\multirow{8}{*}{}& ZeroQ~\cite{ZeroQ}+IL~\cite{inception} & \ding{56} & 26.35 \\ 
\multirow{8}{*}{}& DSG~\cite{DSG}+IL~\cite{inception} & \ding{56} & 43.42 \\
\multirow{8}{*}{}& IntraQ~\cite{IntraQ} & \ding{56} & 48.25 \\ 
\multirow{8}{*}{}& HAST(Ours) & \ding{56} & \textbf{55.67}$\pm0.26$ \\ \hline
\multicolumn{4}{c}{(b) Results on CIFAR-100.} \\ 
\end{tabular} 
}
\vspace{-3mm}
\caption{Results of ResNet-20 on CIFAR-10/100. WBAB indicates the weights and activations are quantized to B-bit.}
\vspace{-10mm}
\label{table1} 
\end{center}   
\end{table}

\subsection{Experimental Setup}
\label{Experimental Setup}
\par \textbf{Datasets.} We evaluate our method on three datasets, including CIFAR-10~\cite{cifar}, CIFAR-100~\cite{cifar} and ImageNet (ILSVRC2012)~\cite{imagenet}, which are commonly used in most ZSQ methods. We report top-1 accuracy of all methods. 
\par \textbf{Network Architectures.} We choose ResNet-20~\cite{resnet} for Cifar-10/100 and select ResNet-18~\cite{resnet}, MobileNetV1~\cite{mobilenetv1} and MobileNetV2~\cite{mobilenetv2} for ImageNet to evaluate our method.
ResNet-20 and ResNet-18 are popular medium-sized models. MobileNetV1 and MobileNetV2 are widely-used lightweight models. All pretrained models are from pytorchcv library and all experiments are implemented with Pytorch~\cite{pytorch}.
\par \textbf{Baselines.} The compared methods based on noise optimization include ZeroQ~\cite{ZeroQ}, DSG~\cite{DSG}, and IntraQ~\cite{IntraQ}.
The generator-based methods are also compared, including GDFQ~\cite{GDFQ}, and AIT~\cite{AIT}.
In addition, SQuant~\cite{SQuant}, an recent on-the-fly ZSQ method without synthesizing data, is also involved. For ZeroQ and DSG, we report the results with the inception loss (IL)~\cite{inception}. Since AIT applies the method on three different generator-based methods, we report the best result. Note all the methods quantize all layers of the models to the ultra-low precisions, except SQuant which sets the last layer to 8-bit.
\par \textbf{Implementation details.} For data generation, the synthetic images are optimized by the loss function Eq.~(\ref{HFNL}) using Adam optimizer with a momentum of 0.9 and the initial learning rate of 0.5. We update the synthetic images for 1,000 iterations and decay the learning rate by 0.1 each time the loss of Eq.~(\ref{HFNL}) stops decreasing for 50 iterations. For all datasets, the batch size is set to 256. We synthesize 5120 images following the settings of IntraQ~\cite{IntraQ}.
For network fine-tuning, the quantized models are fine-tuned by the loss function Eq.~(\ref{eq:FAR}) using SGD with Nesterov with a momentum of 0.9 and the weight decay of $10^{-4}$. The batch size is 256 for CIFAR-10/100 and 16 for ImageNet. The initial learning rate is set to $10^{-5}$ and $10^{-6}$ for CIFAR-10/100 and ImageNet respectively. Both learning rates are decayed by 0.1 every 100 fine-tuning epochs and a total of 150 epochs are given.
For data augmentation and hyper-parameters $\beta$ in Eq.~(\ref{HFNL}) and $\alpha$ in Eq.~(\ref{eq:FAR}), we keep the same settings as~\cite{IntraQ} for fair comparison. In addition, there are three hyper-parameters in our method, including $\gamma$ in Eq.~(\ref{hil}), $\epsilon$ in Eq.~(\ref{SDP}) and $\lambda$ in Eq.~(\ref{eq:FAR}). They are respectively set to 2, 0.01, $1\times10^3$ for CIFAR-10; 2, 0.02, $2\times10^{3}$ for CIFAR-100 and 0.5, 0.01, $4\times10^{3}$ for ImageNet.

\begin{table}[t]   
\begin{center}   
\resizebox{0.48\textwidth}{!}{
\begin{tabular}{c|c|c|c}   
\hline 
Bit-width & Method & Generator & Acc.(\%) \\ \hline \hline
 & full-precision & - & 71.47 \\ \hline
\multirow{8}{*}{W4A4}& Real Data & - & 67.89 \\ \cline{2-4} 
\multirow{8}{*}{}& SQuant~\cite{ZeroQ} & - & 66.14 \\
\multirow{8}{*}{}& GDFQ~\cite{GDFQ} & \ding{52} & 60.60 \\
\multirow{8}{*}{}& AIT~\cite{AIT} & \ding{52} & 66.83 \\
\multirow{8}{*}{}& ZeroQ~\cite{ZeroQ}+IL~\cite{inception} & \ding{56} & 63.38 \\ 
\multirow{8}{*}{}& DSG~\cite{DSG}+IL~\cite{inception} & \ding{56} & 63.11 \\
\multirow{8}{*}{}& IntraQ~\cite{IntraQ} & \ding{56} & 66.47 \\
\multirow{8}{*}{}& HAST(Ours) & \ding{56} & \textbf{66.91}$\pm0.16$ \\ \hline

\multirow{7}{*}{W3A3}& Real Data & - & 51.95 \\ \cline{2-4} 
\multirow{7}{*}{}& SQuant~\cite{ZeroQ} & - & 25.74 \\
\multirow{7}{*}{}& GDFQ~\cite{GDFQ} & \ding{52} & 20.69 \\
\multirow{7}{*}{}& AIT~\cite{AIT} & \ding{52} & 36.70 \\
\multirow{7}{*}{}& ZeroQ~\cite{ZeroQ}+IL~\cite{inception} & \ding{56} & 44.68 \\
\multirow{7}{*}{}& IntraQ~\cite{IntraQ} & \ding{56} & 45.51 \\ 
\multirow{7}{*}{}& HAST(Ours) & \ding{56} & \textbf{51.15}$\pm0.27$ \\ \hline
\end{tabular} 
}
\vspace{-1mm}
\caption{Results of ResNet-18 on ImageNet. WBAB indicates the weights and activations are quantized to B-bit.}
\vspace{-10mm}
\label{table2} 
\end{center}   
\end{table}

\subsection{Comparison Results on CIFAR-10/100}
\label{sec:cifar}

\par In this section, We compare the performance on CIFAR-10/100 against existing ZSQ methods. From Table~\ref{table1}, our method achieves significant performance improvements on both CIFAR-10 and CIFAR-100. Specifically, compared to the advanced generator-based AIT, our method increases the top-1 accuracy of 3-bit quantized models by 7.85\% on CIFAR-10 and 7.03\% on CIFAR-100. When compared with IntraQ which obtains 3-bit accuracy of 77.07\% and 48.25\% on CIFAR-10 and CIFAR-100 respectively by exploiting the intra-class heterogeneity in the synthetic images, the proposed method reaches the higher performance of 88.34\% and 55.67\% using the same number of synthetic images. Notably, 
since the CIFAR-10 dataset is relatively easy when compared with CIFAR-100 and ImageNet, our method outperforms fine-tuning with real data. The gap is also only 0.59\% when it comes to CIFAR-100. 
We obtain the similar results in 4-bit quantization, demonstrating the effectiveness of our proposed HAST.

\begin{table}[t]
\begin{center}
\resizebox{0.48\textwidth}{!}{
\begin{tabular}{c|c|c|c}   
\hline 
Bit-width & Method & Generator & Acc.(\%) \\ \hline \hline
 & full-precision & - & 73.39 \\ \hline
\multirow{8}{*}{W5A5}& Real Data & - & 69.87 \\ \cline{2-4} 
\multirow{8}{*}{}& SQuant~\cite{ZeroQ} & - & 64.20\\
\multirow{8}{*}{}& GDFQ~\cite{GDFQ} & \ding{52} & 59.76 \\
\multirow{8}{*}{}& AIT~\cite{AIT} & \ding{52} & - \\
\multirow{8}{*}{}& ZeroQ~\cite{ZeroQ}+IL~\cite{inception} & \ding{56} & 67.11 \\ 
\multirow{8}{*}{}& DSG~\cite{DSG}+IL~\cite{inception} & \ding{56} & 66.61 \\
\multirow{8}{*}{}& IntraQ~\cite{IntraQ} & \ding{56} & 68.17 \\
\multirow{8}{*}{}& HAST(Ours) & \ding{56} & \textbf{68.52}$\pm0.17$ \\ \hline

\multirow{8}{*}{W4A4}& Real Data & - & 59.66 \\ \cline{2-4} 
\multirow{8}{*}{}& SQuant~\cite{ZeroQ} & - & 10.32 \\
\multirow{8}{*}{}& GDFQ~\cite{GDFQ} & \ding{52} & 28.64 \\
\multirow{8}{*}{}& AIT~\cite{AIT} & \ding{52} & - \\
\multirow{8}{*}{}& ZeroQ~\cite{ZeroQ}+IL~\cite{inception} & \ding{56} & 25.43 \\ 
\multirow{8}{*}{}& DSG~\cite{DSG}+IL~\cite{inception} & \ding{56} & 42.19 \\
\multirow{8}{*}{}& IntraQ~\cite{IntraQ} & \ding{56} & 51.36 \\ 
\multirow{8}{*}{}& HAST(Ours) & \ding{56} & \textbf{57.70}$\pm0.31$ \\ \hline
\multicolumn{4}{c}{(a) Results of MobileNetV1.} \\
\multicolumn{4}{c}{  } \\ \hline
Bit-width & Method & Generator & Acc.(\%) \\ \hline \hline
 & full-precision & & 73.03 \\ \hline
\multirow{8}{*}{W5A5}& Real Data & - & 72.01 \\ \cline{2-4} 
\multirow{8}{*}{}& SQuant~\cite{ZeroQ} & - & 66.83 \\
\multirow{8}{*}{}& GDFQ~\cite{GDFQ} & \ding{52} & 68.14 \\
\multirow{8}{*}{}& AIT~\cite{AIT} & \ding{52} & \textbf{71.96} \\
\multirow{8}{*}{}& ZeroQ~\cite{ZeroQ}+IL~\cite{inception} & \ding{56} & 70.95 \\ 
\multirow{8}{*}{}& DSG~\cite{DSG}+IL~\cite{inception} & \ding{56} & 70.87 \\
\multirow{8}{*}{}& IntraQ~\cite{IntraQ} & \ding{56} & 71.28 \\
\multirow{8}{*}{}& HAST(Ours) & \ding{56} & 71.72$\pm0.19$ \\ \hline

\multirow{8}{*}{W4A4}& Real Data & - & 67.90 \\ \cline{2-4} 
\multirow{8}{*}{}& SQuant~\cite{ZeroQ} & - & 22.07 \\
\multirow{8}{*}{}& GDFQ~\cite{GDFQ} & \ding{52} & 51.30 \\
\multirow{8}{*}{}& AIT~\cite{AIT} & \ding{52} & \textbf{66.47} \\
\multirow{8}{*}{}& ZeroQ~\cite{ZeroQ}+IL~\cite{inception} & \ding{56} & 60.15 \\ 
\multirow{8}{*}{}& DSG~\cite{DSG}+IL~\cite{inception} & \ding{56} & 59.04 \\
\multirow{8}{*}{}& IntraQ~\cite{IntraQ} & \ding{56} & 65.10 \\ 
\multirow{8}{*}{}& HAST(Ours) & \ding{56} & 65.60$\pm0.27$ \\ \hline
\multicolumn{4}{c}{(b) Results of MobileNetV2.} \\ 
\end{tabular} 
}
\vspace{-3mm}
\caption{Results of MobileNetV1/V2 on ImageNet. WBAB indicates the weights and activations are quantized to B-bit}
\vspace{-10mm}
\label{table3} 
\end{center}   
\end{table}

\subsection{Comparison Results on ImageNet}
\label{sec:imagenet}
\par We further compare with the competitors on the large-scale ImageNet. The quantized networks include ResNet-18 and MobileNetV1/V2. Similar to CIFAR-10/100, we quantize all layers of the networks. Differently, since MobileNetV1/V2 are lightweight models which suffer great performance degradation when quantized to ultra-low precision, we quantize MobileNetV1/V2 to 5-bit and 4-bit following the settings of most existing methods.
\par \textbf{ResNet-18.} Table~\ref{table2} shows the experimental results of ResNet-18. In the case of 4-bit, our HSAT (66.91\%) slightly outperforms the IntraQ (66.47\%) and AIT (66.83\%). When it comes to 3-bit, our HSAT significantly surpasses AIT and IntraQ by 14.45\% and 5.64\% on ImageNet. Moreover, the performance of our method is very close to fine-tuning using real data regardless of 4-bit or 3-bit, with a gap of less than 1\%. This indicates that our HSAT 
is able to mimic real data and generalize quantized model well even on large-scale and hard datasets.


\par \textbf{MobileNetV1/V2.} In Table~\ref{table3}, our method still outperforms almost all baselines except AIT on quantizing lightweight models. For example, while other methods achieve unexpectedly low performance on quantizing MobileNetV1 to 4-bit, our HSAT obtains 6.34\% accuracy improvement when compared with IntraQ. However, Unlike other networks, we find that the fine-tuning process of MobileNetV2 is more unstable and the performance improvement during fine-tuning is smaller. This shows that the training process of MobileNetV2 has a greater impact on the performance recovery than the sample difficulty. AIT uses a dynamic learning rate for each convolution kernel to solve this problem, while we use a uniform fixed learning rate. Thus AIT performs better than HAST on MobileNetV2. It is worth mentioning that AIT can be used in combination with HAST as long as memory and time overheads allow.

\par In short, sufficient experiments over various network architectures demonstrate that the synthetic data generated by the proposed HSAT scheme is able to significantly improve the performance of the quantized model. The results also suggest that synthesizing and learning hard samples is important for improving the quantized model. Especially when model is quantized to lower bit-width, the effect of hard samples is more obvious on final performance.

\subsection{Ablation Study}
\label{sec:ablation}
\par In this section, we conduct ablation studies of the hyper-parameters and investigate the effect of the three components of our method. We put the hyper-parameters ablation studies, further discussions and related experimental results in the appendix.


\par \textbf{Effects of the proposed components.} We further study the effectiveness of our proposed hard sample synthesis in Sec.~\ref{HSS}, sample difficulty promotion in Sec.~\ref{sec:SDP}, and feature alignment in Sec.~\ref{FA}. Table~\ref{table4} shows the experimental results. Note that when we use ZeroQ+IL as the baseline described in Sec.~\ref{Preliminaries}, we only obtain an accuracy of 44.68\%. From Table~\ref{table4}, when the three components are individually added to synthesize and learn hard samples, the accuracy increases compared with the baseline. Among them, adversarial augmentation obtains a high accuracy improvement of 3.79\%. This inspires us the importance of hard samples in training quantized models. Since attention transfer can help the quantized model learn more informative knowledge from the intermediate feature of the full-precision model, it still increases the performance by 1.39\% without hard samples. Furthermore, the performance continues to increase when any two of them are used. When all of them are applied, we obtain the best performance of 51.15\%.

\begin{table}[h]   
\begin{center}
\begin{tabular}{c c c|c}   
\hline 
HSS & SDP & FA & Acc.(\%) \\ \hline \hline
 & & & 44.68\\ \hline
\ding{52} & & & 45.56\\
 & \ding{52} & & 48.47\\
 & & \ding{52} & 46.07\\
 & \ding{52} & \ding{52} & 50.22\\
\ding{52} & \ding{52} & & 49.17\\
\ding{52} & & \ding{52} & 47.94\\
\ding{52} & \ding{52} & \ding{52} & \textbf{51.15}\\ \hline
\end{tabular} 
\vspace{-1mm}
\caption{Ablations on different components of our method. “HSS” indicates the hard sample synthesis, “SDP” indicates the sample difficulty promotion, and “FA” indicates the feature alignment. We report the top-1 accuracy of 3-bit ResNet-18 on ImageNet.}
\vspace{-7mm}
\label{table4} 
\end{center}   
\end{table}

\par \textbf{Limitation.} In this work, we merely considered a limited number of scenarios. Thus, we will explore the potential power for more practical scenarios, such as federated learning~\cite{Dong_2022_CVPR,tang2022virtual} (with limited communication bandwidth), vision transformer~\cite{dosovitskiy2020image} (with large-scale model parameters), and the performance of quantized models over out-of-distribution scenarios~\cite{fangout}.

\section{Conclusion}
\label{sec:conclusion}
\par In this paper, we investigate the state-of-the-art solutions based on noise optimization for zero-shot quantization and demonstrate that the synthetic samples are easy for the quantized model to fit, which harms the performance of the quantized model on real test data. Thus, hard samples matter a lot for zero-shot quantization. We achieve the goal by not only paying more attention to generating hard samples but also making samples harder to fit during fine-tuning. Feature alignment is applied in the fine-tuning process to help the quantized model learn hard samples better. Our method can achieve comparable performance with those fine-tuned using real data.

\section{Acknowledgements}
This work was partially supported by the Key-Area Research and Development Program of Guangdong Province 2019B010155002, Science and Technology Program of Guangzhou, China under Grants 202007030007, Program for Guangdong Introducing Innovative and Enterpreneurial Teams 2017ZT07X183. YGZ and BH were supported by NSFC Young Scientists Fund No. 62006202 and Guangdong Basic and Applied Basic Research Foundation No. 2022A1515011652.

{\small
\bibliographystyle{ieee_fullname}
\bibliography{11_references}

\begin{thebibliography}{10}\itemsep=-1pt

\bibitem{DBLP:conf/nips/BannerNS19}
Ron Banner, Yury Nahshan, and Daniel Soudry.
\newblock Post training 4-bit quantization of convolutional networks for
  rapid-deployment.
\newblock In Hanna~M. Wallach, Hugo Larochelle, Alina Beygelzimer, Florence
  d'Alch{\'{e}}{-}Buc, Emily~B. Fox, and Roman Garnett, editors, {\em NeurIPS},
  2019.

\bibitem{ZeroQ}
Yaohui Cai, Zhewei Yao, Zhen Dong, Amir Gholami, Michael~W. Mahoney, and Kurt
  Keutzer.
\newblock Zeroq: {A} novel zero shot quantization framework.
\newblock In {\em CVPR}, 2020.

\bibitem{chen2022active}
Peihao Chen, Dongyu Ji, Kunyang Lin, Weiwen Hu, Wenbing Huang, Thomas~H. Li,
  Mingkui Tan, and Chuang Gan.
\newblock Learning active camera for multi-object navigation.
\newblock In {\em Neural Information Processing Systems (NeurIPS)}, 2022.

\bibitem{chen2022weakly}
Peihao Chen, Dongyu Ji, Kunyang Lin, Runhao Zeng, Thomas~H Li, Mingkui Tan, and
  Chuang Gan.
\newblock Weakly-supervised multi-granularity map learning for
  vision-and-language navigation.
\newblock {\em NeurIPS}, 2022.

\bibitem{PACT}
Jungwook Choi, Zhuo Wang, Swagath Venkataramani, Pierce~I{-}Jen Chuang,
  Vijayalakshmi Srinivasan, and Kailash Gopalakrishnan.
\newblock {PACT:} parameterized clipping activation for quantized neural
  networks.
\newblock {\em arXiv preprint arXiv:1805.06085}, 2018.

\bibitem{Qimera}
Kanghyun Choi, Deokki Hong, Noseong Park, Youngsok Kim, and Jinho Lee.
\newblock Qimera: Data-free quantization with synthetic boundary supporting
  samples.
\newblock In {\em NeurIPS}, 2021.

\bibitem{AIT}
Kanghyun Choi, Hyeyoon Lee, Deokki Hong, Joonsang Yu, Noseong Park, Youngsok
  Kim, and Jinho Lee.
\newblock It's all in the teacher: Zero-shot quantization brought closer to the
  teacher.
\newblock In {\em CVPR}, 2022.

\bibitem{What_Transferred_Dong_CVPR2020}
Jiahua Dong, Yang Cong, Gan Sun, Bineng Zhong, and Xiaowei Xu.
\newblock What can be transferred: Unsupervised domain adaptation for
  endoscopic lesions segmentation.
\newblock In {\em CVPR}, pages 4022--4031, June 2020.

\bibitem{Dong_2022_CVPR}
Jiahua Dong, Lixu Wang, Zhen Fang, Gan Sun, Shichao Xu, Xiao Wang, and Qi Zhu.
\newblock Federated class-incremental learning.
\newblock In {\em CVPR}, pages 10164--10173, June 2022.

\bibitem{dosovitskiy2020image}
Alexey Dosovitskiy, Lucas Beyer, Alexander Kolesnikov, Dirk Weissenborn,
  Xiaohua Zhai, Thomas Unterthiner, Mostafa Dehghani, Matthias Minderer, Georg
  Heigold, Sylvain Gelly, et~al.
\newblock An image is worth 16x16 words: Transformers for image recognition at
  scale.
\newblock {\em ICLR}, 2021.

\bibitem{DBLP:journals/tetci/DuanYTZT22}
Jiafei Duan, Samson Yu, Hui~Li Tan, Hongyuan Zhu, and Cheston Tan.
\newblock A survey of embodied {AI:} from simulators to research tasks.
\newblock {\em {IEEE} Trans. Emerg. Top. Comput. Intell.}, 6(2):230--244, 2022.

\bibitem{LSQ}
Steven~K. Esser, Jeffrey~L. McKinstry, Deepika Bablani, Rathinakumar Appuswamy,
  and Dharmendra~S. Modha.
\newblock Learned step size quantization.
\newblock In {\em ICLR}, 2020.

\bibitem{DBLP:journals/ijcv/EveringhamEGWWZ15}
Mark Everingham, S.~M.~Ali Eslami, Luc~Van Gool, Christopher K.~I. Williams,
  John~M. Winn, and Andrew Zisserman.
\newblock The pascal visual object classes challenge: {A} retrospective.
\newblock {\em International Journal of Computer Vision}, 111(1):98--136, 2015.

\bibitem{fangout}
Zhen Fang, Yixuan Li, Jie Lu, Jiahua Dong, Bo Han, and Feng Liu.
\newblock Is out-of-distribution detection learnable?
\newblock In {\em NeurIPS}, 2022.

\bibitem{DBLP:conf/iccv/Girshick15}
Ross~B. Girshick.
\newblock Fast {R-CNN}.
\newblock In {\em ICCV}, 2015.

\bibitem{DBLP:conf/cvpr/GirshickDDM14}
Ross~B. Girshick, Jeff Donahue, Trevor Darrell, and Jitendra Malik.
\newblock Rich feature hierarchies for accurate object detection and semantic
  segmentation.
\newblock In {\em CVPR}, 2014.

\bibitem{DSQ}
Ruihao Gong, Xianglong Liu, Shenghu Jiang, Tianxiang Li, Peng Hu, Jiazhen Lin,
  Fengwei Yu, and Junjie Yan.
\newblock Differentiable soft quantization: Bridging full-precision and low-bit
  neural networks.
\newblock In {\em CVPR}, 2019.

\bibitem{FGSM}
Ian~J. Goodfellow, Jonathon Shlens, and Christian Szegedy.
\newblock Explaining and harnessing adversarial examples.
\newblock In {\em ICLR}, 2015.

\bibitem{SQuant}
Cong Guo, Yuxian Qiu, Jingwen Leng, Xiaotian Gao, Chen Zhang, Yunxin Liu, Fan
  Yang, Yuhao Zhu, and Minyi Guo.
\newblock Squant: On-the-fly data-free quantization via diagonal hessian
  approximation.
\newblock In {\em ICLR}, 2022.

\bibitem{inception}
Matan Haroush, Itay Hubara, Elad Hoffer, and Daniel Soudry.
\newblock The knowledge within: Methods for data-free model compression.
\newblock In {\em CVPR}, 2020.

\bibitem{resnet}
Kaiming He, Xiangyu Zhang, Shaoqing Ren, and Jian Sun.
\newblock Deep residual learning for image recognition.
\newblock In {\em CVPR}, 2016.

\bibitem{GZNQ}
Xiangyu He, Jiahao Lu, Weixiang Xu, Qinghao Hu, Peisong Wang, and Jian Cheng.
\newblock Generative zero-shot network quantization.
\newblock In {\em CVPRW}, 2021.

\bibitem{mobilenetv1}
Andrew~G. Howard, Menglong Zhu, Bo Chen, Dmitry Kalenichenko, Weijun Wang,
  Tobias Weyand, Marco Andreetto, and Hartwig Adam.
\newblock Mobilenets: Efficient convolutional neural networks for mobile vision
  applications.
\newblock {\em arXiv preprint arXiv:1704.04861}, 2017.

\bibitem{DBLP:conf/cvpr/JacobKCZTHAK18}
Benoit Jacob, Skirmantas Kligys, Bo Chen, Menglong Zhu, Matthew Tang, Andrew~G.
  Howard, Hartwig Adam, and Dmitry Kalenichenko.
\newblock Quantization and training of neural networks for efficient
  integer-arithmetic-only inference.
\newblock In {\em CVPR}, 2018.

\bibitem{QIL}
Sangil Jung, Changyong Son, Seohyung Lee, JinWoo Son, Jae{-}Joon Han, Youngjun
  Kwak, Sung~Ju Hwang, and Changkyu Choi.
\newblock Learning to quantize deep networks by optimizing quantization
  intervals with task loss.
\newblock In {\em CVPR}, 2019.

\bibitem{cifar}
Alex Krizhevsky and Geoffrey Hinton.
\newblock Learning multiple layers of features from tiny images.
\newblock {\em University of Toronto}, 2009.

\bibitem{imagenet}
Alex Krizhevsky, Ilya Sutskever, and Geoffrey~E. Hinton.
\newblock Imagenet classification with deep convolutional neural networks.
\newblock In {\em NeurIPS}, 2012.

\bibitem{DBLP:journals/cacm/KrizhevskySH17}
Alex Krizhevsky, Ilya Sutskever, and Geoffrey~E. Hinton.
\newblock Imagenet classification with deep convolutional neural networks.
\newblock {\em Communications of the ACM}, 60(6):84--90, 2017.

\bibitem{DBLP:conf/cvpr/Lee0H21}
Junghyup Lee, Dohyung Kim, and Bumsub Ham.
\newblock Network quantization with element-wise gradient scaling.
\newblock In {\em CVPR}, 2021.

\bibitem{GHM}
Buyu Li, Yu Liu, and Xiaogang Wang.
\newblock Gradient harmonized single-stage detector.
\newblock In {\em AAAI}, 2019.

\bibitem{APoT}
Yuhang Li, Xin Dong, and Wei Wang.
\newblock Additive powers-of-two quantization: An efficient non-uniform
  discretization for neural networks.
\newblock In {\em ICLR}, 2020.

\bibitem{BRECQ}
Yuhang Li, Ruihao Gong, Xu Tan, Yang Yang, Peng Hu, Qi Zhang, Fengwei Yu, Wei
  Wang, and Shi Gu.
\newblock {BRECQ:} pushing the limit of post-training quantization by block
  reconstruction.
\newblock In {\em ICLR}, 2021.

\bibitem{MixMix}
Yuhang Li, Feng Zhu, Ruihao Gong, Mingzhu Shen, Xin Dong, Fengwei Yu, Shaoqing
  Lu, and Shi Gu.
\newblock Mixmix: All you need for data-free compression are feature and data
  mixing.
\newblock In {\em ICCV}, 2021.

\bibitem{DBLP:conf/nips/LinJX00WHL20}
Mingbao Lin, Rongrong Ji, Zihan Xu, Baochang Zhang, Yan Wang, Yongjian Wu,
  Feiyue Huang, and Chia{-}Wen Lin.
\newblock Rotated binary neural network.
\newblock In {\em NeurIPS}, 2020.

\bibitem{ZAQ}
Yuang Liu, Wei Zhang, and Jun Wang.
\newblock Zero-shot adversarial quantization.
\newblock In {\em CVPR}, 2021.

\bibitem{Survey}
Yuang Liu, Wei Zhang, Jun Wang, and Jianyong Wang.
\newblock Data-free knowledge transfer: {A} survey.
\newblock {\em arXiv preprint arXiv:2112.15278}, 2021.

\bibitem{DBLP:conf/iclr/MartinezYBT20}
Brais Mart{\'{\i}}nez, Jing Yang, Adrian Bulat, and Georgios Tzimiropoulos.
\newblock Training binary neural networks with real-to-binary convolutions.
\newblock In {\em ICLR}, 2020.

\bibitem{DFQ}
Markus Nagel, Mart van Baalen, Tijmen Blankevoort, and Max Welling.
\newblock Data-free quantization through weight equalization and bias
  correction.
\newblock In {\em ICCV}, 2019.

\bibitem{DBLP:conf/cvpr/PangCSFOL19}
Jiangmiao Pang, Kai Chen, Jianping Shi, Huajun Feng, Wanli Ouyang, and Dahua
  Lin.
\newblock Libra {R-CNN:} towards balanced learning for object detection.
\newblock In {\em CVPR}, 2019.

\bibitem{pytorch}
Adam Paszke, Sam Gross, Francisco Massa, and Adam Lerer.
\newblock Pytorch: An imperative style, high-performance deep learning library.
\newblock In {\em NeurIPS}, 2019.

\bibitem{DBLP:conf/cvpr/QinGLSWYS20}
Haotong Qin, Ruihao Gong, Xianglong Liu, Mingzhu Shen, Ziran Wei, Fengwei Yu,
  and Jingkuan Song.
\newblock Forward and backward information retention for accurate binary neural
  networks.
\newblock In {\em CVPR}, 2020.

\bibitem{mobilenetv2}
Mark Sandler, Andrew~G. Howard, Menglong Zhu, Andrey Zhmoginov, and
  Liang{-}Chieh Chen.
\newblock Mobilenetv2: Inverted residuals and linear bottlenecks.
\newblock In {\em CVPR}, 2018.

\bibitem{DBLP:journals/corr/SimonyanZ14a}
Karen Simonyan and Andrew Zisserman.
\newblock Very deep convolutional networks for large-scale image recognition.
\newblock In {\em ICLR}, 2015.

\bibitem{DBLP:conf/cvpr/SzegedyLJSRAEVR15}
Christian Szegedy, Wei Liu, Yangqing Jia, Pierre Sermanet, Scott~E. Reed,
  Dragomir Anguelov, Dumitru Erhan, Vincent Vanhoucke, and Andrew Rabinovich.
\newblock Going deeper with convolutions.
\newblock In {\em CVPR}, 2015.

\bibitem{tang2022virtual}
Zhenheng Tang, Yonggang Zhang, Shaohuai Shi, Xin He, Bo Han, and Xiaowen Chu.
\newblock Virtual homogeneity learning: Defending against data heterogeneity in
  federated learning.
\newblock In {\em International Conference on Machine Learning}, pages
  21111--21132. PMLR, 2022.

\bibitem{GDFQ}
Shoukai Xu, Haokun Li, Bohan Zhuang, Jing Liu, Jiezhang Cao, Chuangrun Liang,
  and Mingkui Tan.
\newblock Generative low-bitwidth data free quantization.
\newblock In {\em ECCV}, 2020.

\bibitem{PyHessian}
Zhewei Yao, Amir Gholami, Kurt Keutzer, and Michael~W. Mahoney.
\newblock Pyhessian: Neural networks through the lens of the hessian.
\newblock In {\em IEEE International Conference on Big Data}, 2020.

\bibitem{featurealignment}
Jason Yosinski, Jeff Clune, Yoshua Bengio, and Hod Lipson.
\newblock How transferable are features in deep neural networks?
\newblock In {\em NeurIPS}, 2014.

\bibitem{AttentionTransfer}
Sergey Zagoruyko and Nikos Komodakis.
\newblock Paying more attention to attention: Improving the performance of
  convolutional neural networks via attention transfer.
\newblock In {\em ICLR}, 2017.

\bibitem{DSG}
Xiangguo Zhang, Haotong Qin, Yifu Ding, Ruihao Gong, Qinghua Yan, Renshuai Tao,
  Yuhang Li, Fengwei Yu, and Xianglong Liu.
\newblock Diversifying sample generation for accurate data-free quantization.
\newblock In {\em CVPR}, 2021.

\bibitem{zhang2021causaladv}
Yonggang Zhang, Mingming Gong, Tongliang Liu, Gang Niu, Xinmei Tian, Bo Han,
  Bernhard Sch{\"o}lkopf, and Kun Zhang.
\newblock Causaladv: Adversarial robustness through the lens of causality.
\newblock 2022.

\bibitem{IntraQ}
Yunshan Zhong, Mingbao Lin, Gongrui Nan, Jianzhuang Liu, Baochang Zhang,
  Yonghong Tian, and Rongrong Ji.
\newblock Intraq: Learning synthetic images with intra-class heterogeneity for
  zero-shot network quantization.
\newblock In {\em CVPR}, 2022.

\bibitem{AutoReCon}
Baozhou Zhu, H.~Peter Hofstee, Johan Peltenburg, Jinho Lee, and Zaid Al{-}Ars.
\newblock Autorecon: Neural architecture search-based reconstruction for
  data-free compression.
\newblock In {\em IJCAI}, 2021.

\bibitem{DBLP:conf/cvpr/ZhuangLTSR20}
Bohan Zhuang, Lingqiao Liu, Mingkui Tan, Chunhua Shen, and Ian~D. Reid.
\newblock Training quantized neural networks with a full-precision auxiliary
  module.
\newblock In {\em CVPR}, 2020.

\bibitem{DBLP:conf/cvpr/ZhuangSTL019}
Bohan Zhuang, Chunhua Shen, Mingkui Tan, Lingqiao Liu, and Ian~D. Reid.
\newblock Structured binary neural networks for accurate image classification
  and semantic segmentation.
\newblock In {\em CVPR}, 2019.

\end{thebibliography}
}

\twocolumn[
\begin{@twocolumnfalse}
	\section*{\centering \Large{Supplementary Material for \\ \emph{Hard Sample Matters a Lot in Zero-Shot Quantization\\[50pt]}}}
\end{@twocolumnfalse}
]

\appendix
\label{sec:appendix}

\section{Sensitivity Analysis on Hyper-parameters}
\par
\begin{figure*}[ht]
\centering
\begin{minipage}[b]{\linewidth}
        \centering
        \includegraphics[width=2.1in]{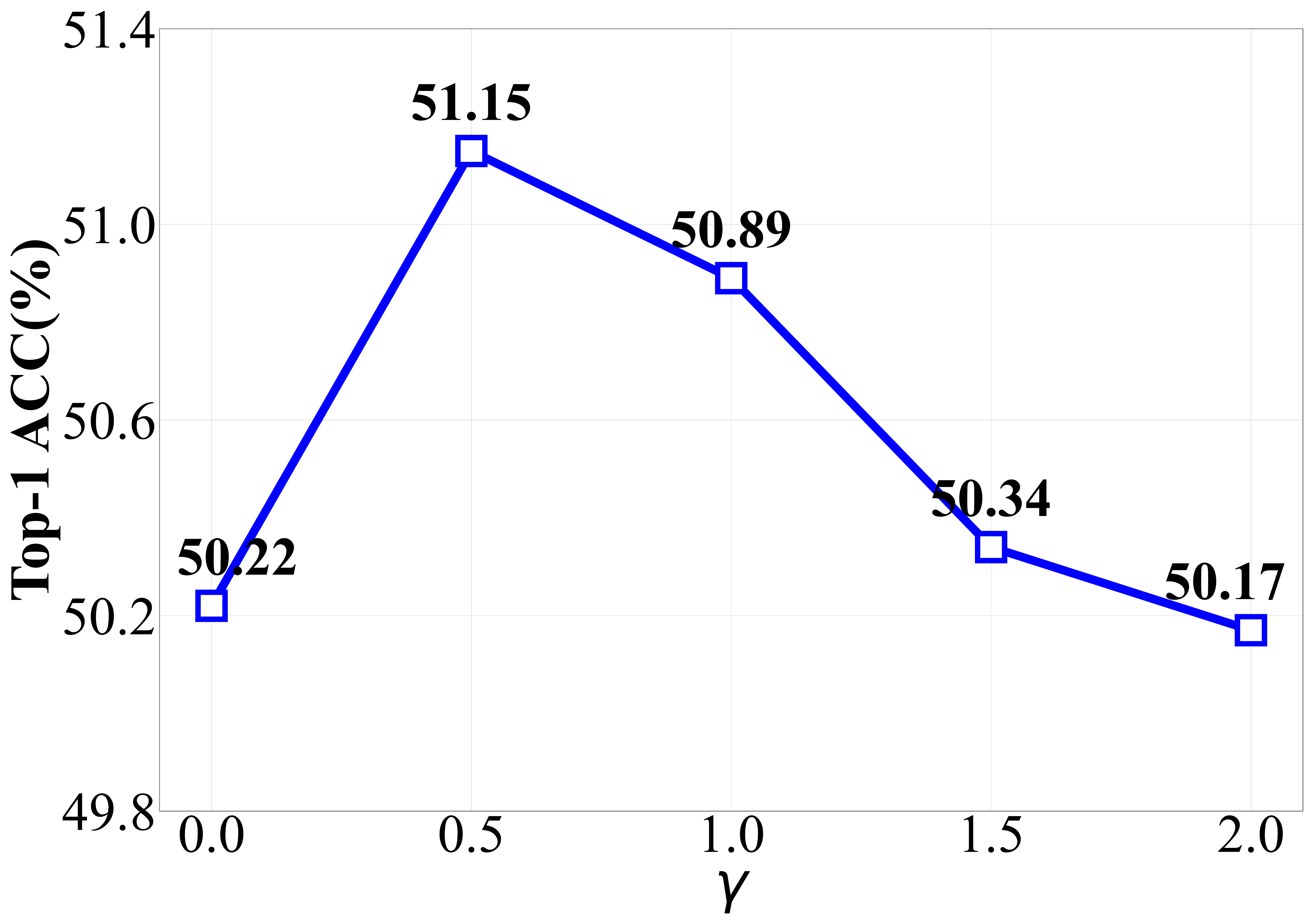}
        \includegraphics[width=2.1in]{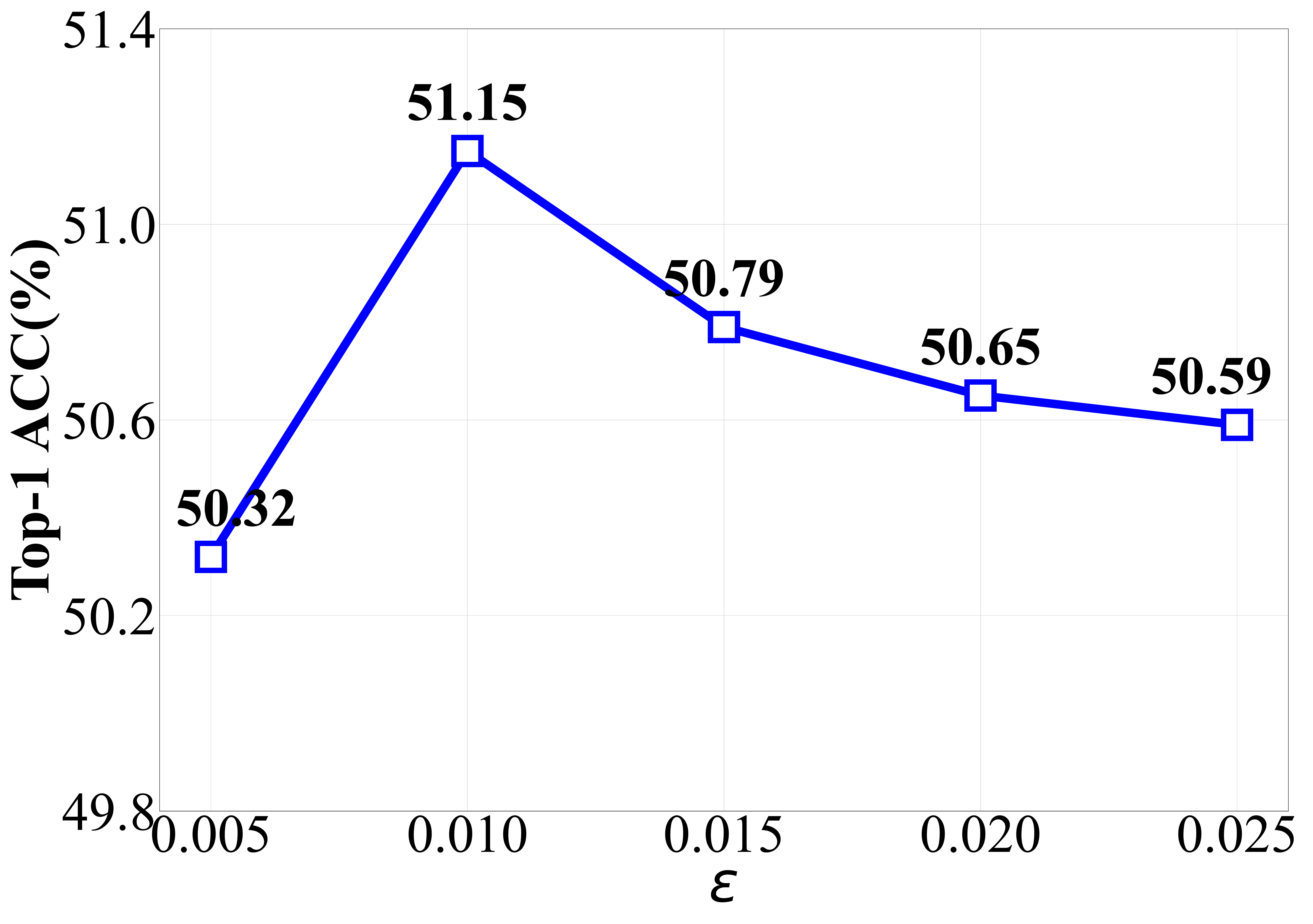}
        \includegraphics[width=2.1in]{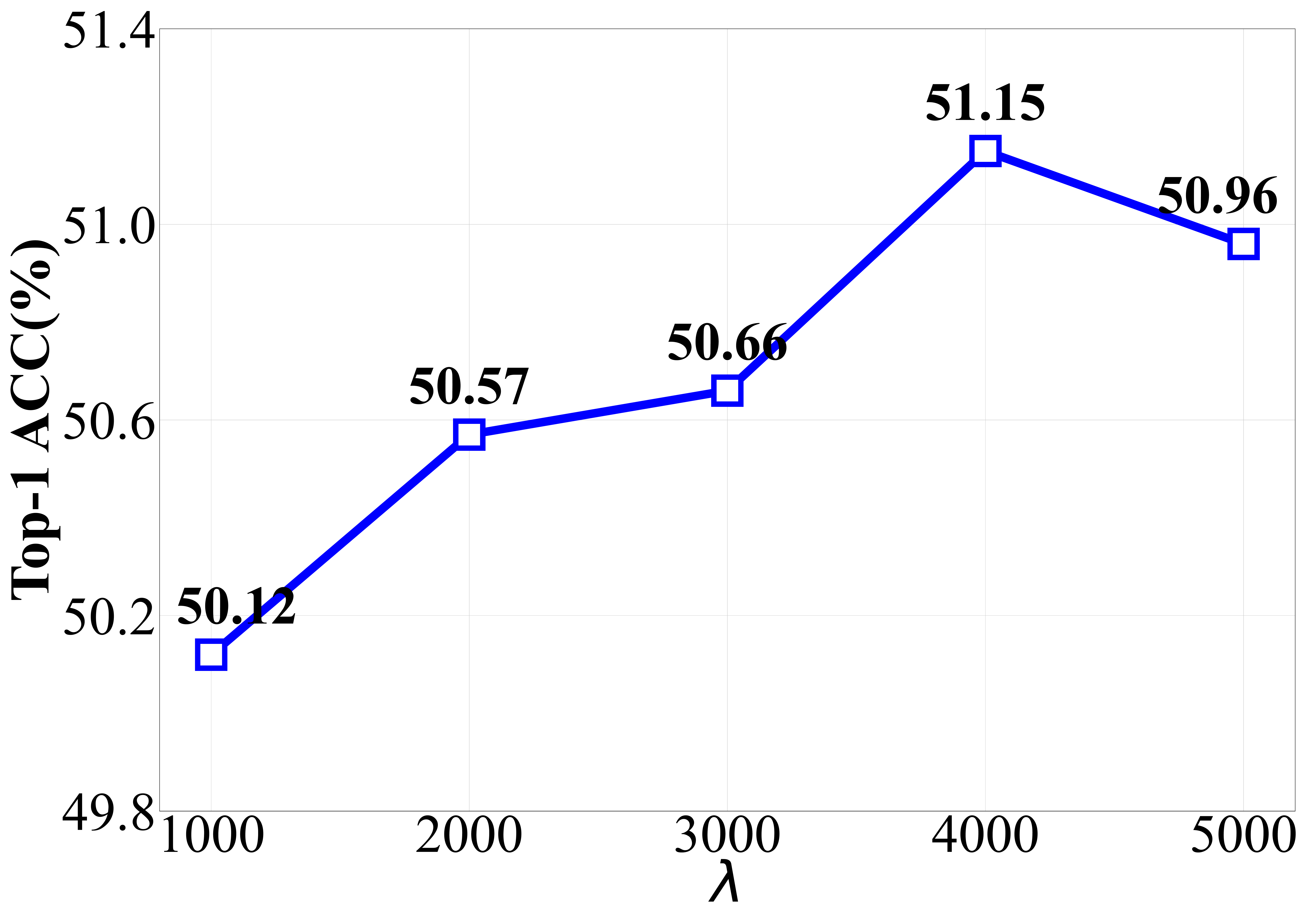}
\end{minipage}
\vspace{-8mm}
\caption{Sensitivity analysis on hyper-parameters. We report the top-1 accuracy of 3-bit ResNet-18 on ImageNet.}
\label{supp_fig1}
\end{figure*}

 The hyper-parameters in our method including $\gamma$ in Eq.~8, $\epsilon$ in Eq.~10 and $\lambda$ in Eq.~12. In Figure~\ref{supp_fig1}, a sensitivity study of them on 3-bit ResNet-18 is performed. For $\gamma$, we choose several small values from 0 to 2 since the optimization in data generation process of ImageNet is not as easy as other small-scale datasets. We keep $\epsilon$ on a small magnitude so that the perturbation is not too large, and keep $\lambda$ on a large magnitude so that the magnitude of two loss terms are consistent. 
The results show that the performance of HAST is somewhat sensitive to these hyper-parameters, but most of these results ($50.12\% \sim 51.15\%$) are comparable with that of the model fine-tuned on real data ($51.95\%$). Note that the worse results in Figure~\ref{supp_fig1} outperforms the quantized model obtained by the state-of-the-art ZSQ method ($45.51\%$) in a large margin.
We conduct similar experiments to find out the optimal value of these hyper-parameters on other datasets.

\section{Sample Difficulty Promotion Details}
\label{SDP Details}
\par
\textbf{Perturbation Direction Calculation.} In the main paper, we calculate the perturbation $\delta$ by maximizing the sample difficulty, which is closely related to the loss. However, there are two loss terms, i.e., the Kullback-Leibler (KL) loss and the feature alignment (FA) loss in the fine-tuning process. Thus we conduct a further experiment to select the optimal loss for perturbation direction calculation. The experimental results are shown in Table~\ref{supp_table1}. We observe that the choice of loss for calculating the perturbation direction has a certain impact on the performance. Though not optimal for all settings, we choose KL+FA to calculate perturbation direction since it shows the best in most settings.

\begin{table}[h]   
\begin{center}   
\resizebox{\columnwidth}{!}{
\begin{tabular}{c c c c c c}   
\hline 
Dataset & Model & Bit-width & KL & FA & KL+FA \\ \hline
\multirow{2}*{Cifar-10}& \multirow{2}*{ResNet-20} & W4A4 & 92.43 & 92.29 & 92.36 \\ 
\multirow{2}*{}& \multirow{2}*{} & W3A3 & 88.29 & 87.68 & 88.34 \\ \hline
\multirow{2}*{Cifar-100}& \multirow{2}*{ResNet-20} & W4A4 & 66.69 & 66.50 & 66.68 \\ 
\multirow{2}*{}& \multirow{2}*{} & W3A3 & 55.61 & 55.13 & 55.67 \\ \hline
\multirow{2}*{ImageNet}& \multirow{2}*{ResNet-18} & W4A4 & 66.90 & 66.69 & 66.91 \\ 
\multirow{2}*{}& \multirow{2}*{} & W3A3 & 51.06 & 50.87 & 51.15  \\ \hline
\end{tabular} 
}
\caption{Performance of our HAST when calculating perturbation direction with diferent losses. We maximize the gradient of KL, FA and KL+FA respectively to calculate perturbation direction.}
\label{supp_table1} 
\end{center}   
\end{table}

\par
\textbf{loss weights.} We apply sample difficulty promotion to the synthetic samples obtained by hard sample synthesis for more difficult samples. Then both of them are used to fine-tune the quantized model with the same loss weights. Further experiments on the loss weights of the original synthetic samples and the promotional samples are conducted. Experimental results are shown in table~\ref{supp_table2}. The loss weight of the original synthetic samples is denoted as $a$, and that of the promotional samples is denoted as $b$. We perform 3-bit quantization on CIFAR-10 and ImageNet. For CIFAR-10, we achieve the best accuracy of 88.34\% by setting both the weights to 1. When it comes to ImageNet, better performance than that reported in the main paper is obtained by increasing the weight of promotional samples.

\begin{table}[ht]   
\begin{center}   
\resizebox{\columnwidth}{!}{
\begin{tabular}{c c c c c c}   
\hline 
\multirow{2}{*}{$a,b$} & ResNet-20 & ResNet-18 & \multirow{2}{*}{$a,b$} & ResNet-20 & ResNet-18  \\ \cline{2-3} \cline{5-6}
 & Cifar-10 & ImageNet &  & Cifar-10 & ImageNet  \\ \hline
 1,0 & 86.17 & 47.94 & 0,1 & 88.19 & 48.55 \\
 3,1 & 85.92 & 50.52 & 1,4 & 86.69 & 52.14 \\
 2,1 & 87.53 & 50.97 & 1,3 & 86.94 & 53.12 \\
 1,1 & 88.34 & 51.15 & 1,2 & 87.73 & 52.69 \\ \hline
\end{tabular} 
}
\caption{Ablation results of loss weights in W3A3 setting. The loss weights of original synthetic samples and promotional samples are denoted as $a,b$ respectively.}
\label{supp_table2} 
\end{center}   
\end{table}

\section{Feature Alignment Analysis}
\par 
\textbf{Direct feature alignment vs. relaxed feature alignment.} Direct feature alignment~\cite{featurealignment} is an easy and effective way to transfer feature representations by directly using mean square error to align the feature. However, we use attention vector~\cite{AttentionTransfer} to relax the feature alignment constraint due to the limited capacity of quantized model. In this section, we provide the performance comparison of our HAST between using direct feature alignment (DFA) and using relaxed feature alignment (RFA). Table~\ref{supp_table3} shows the experimental results. The relaxed feature alignment obtains better performance in any settings over direct feature alignment. Significant improvements can be observed from 3-bit quantization. This shows that it is harmful for low-precision quantized model to learn the feature representations of full-precision model directly.

\begin{table}[ht]   
\begin{center}   
\resizebox{\columnwidth}{!}{
\begin{tabular}{c c c c c}   
\hline 
Dataset & Model & Bit-width & HAST(DFA) & HAST(RFA) \\ \hline
\multirow{2}*{Cifar-10}& \multirow{2}*{ResNet-20} & W4A4 & 91.99 & 92.36 \\ 
\multirow{2}*{}& \multirow{2}*{} & W3A3 & 83.92 & 88.34 \\ \hline
\multirow{2}*{Cifar-100}& \multirow{2}*{ResNet-20} & W4A4 & 66.53 & 66.68 \\ 
\multirow{2}*{}& \multirow{2}*{} & W3A3 & 51.50 & 55.67 \\ \hline
\multirow{2}*{ImageNet}& \multirow{2}*{ResNet-18} & W4A4 & 66.49 & 66.91 \\ 
\multirow{2}*{}& \multirow{2}*{} & W3A3 & 45.52 & 51.15 \\ \hline
\end{tabular} 
}
\caption{Performance of our HAST with direct feature alignment and relaxed feature alignment.}
\label{supp_table3} 
\end{center}   
\end{table}

\begin{figure*}[ht]
\centering
    \subfloat[Gradient cosine similarity.]{
        \label{supp_fig2.a}
        \includegraphics[width=1.6in]{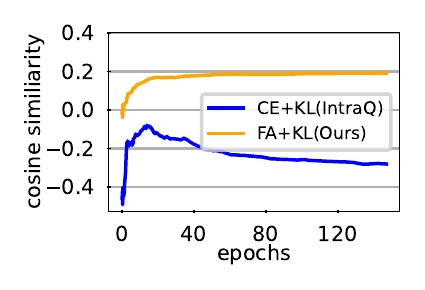}
    }
    \quad
    \subfloat[Distribution of eigenvalues.]{
        \label{supp_fig2.b}
        \includegraphics[width=1.6in]{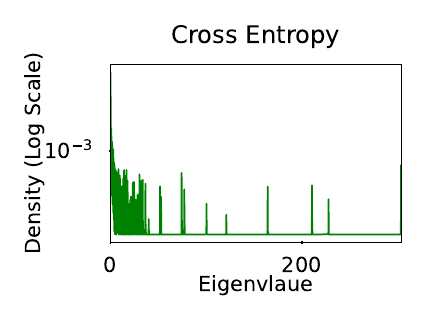}
        \includegraphics[width=1.6in]{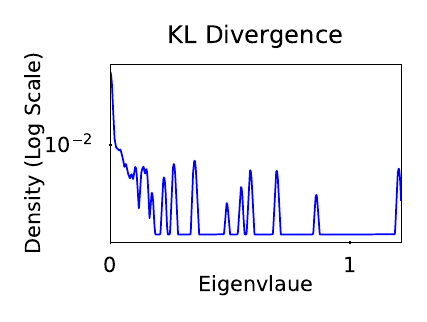}
        \includegraphics[width=1.6in]{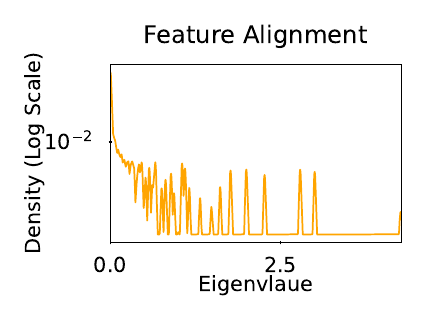}
    }
\caption{Further experiments on feature alignment. (a)Gradient cosine similarity of two terms in loss function. (b)Distribution of the eigenvalues for different loss.}
\label{supp_fig2}
\end{figure*}

\par 
\textbf{Cooperation with KL.} Gradient cosine similarity was used in~\cite{AIT} to measure the cooperation ability of multiple loss terms. The authors found that the cross-entropy (CE) loss does not work well with the Kullback-Leibler (KL) loss in network fine-tuning process. We apply this metric in our work. Specifically, we fine-tune the 3-bit ResNet-20 using baseline (CE+KL)~\cite{IntraQ} and our HAST (FA+KL) respectively and measure the cosine similarity of the gradient of two distinct loss terms. As shown in Figure~\ref{supp_fig2.a}, the cosine distance between CE and KL takes negative values throughout the fine-tuning, while that of FA+KL is positive. This implies that the combinations of FA and KL cooperate well, and using them together could enhance each other, which is opposite to the combinations of CE and KL.

\par \textbf{Generalizability.}  Hessian matrix was used in~\cite{AIT} to measure the local curvature of the loss surface and compare the generalizability of the two distinct loss terms. Since Hessian matrix itself is enormous in size and computations involving its entirety is considered almost infeasible, analyzing the eigenvalues of the matrix is often the most preferred way to study its characteristics. Figure~\ref{supp_fig2.b} plots the distribution of the eigenvalues of the Hessian matrix, approximated by PyHessian~\cite{PyHessian}. We separate Hessian calculation for each loss of CE, KL and FA. A huge difference in the local curvature of the loss terms can be observed. While CE has longer tail for high eigenvalues, KL and FA has more concentration to lower eigenvalues, which means the local curvature of loss surface of KL and FA is smaller than that of CE, leading to better genrealizability according to the finding that  smaller local curvature improves generalization~\cite{AIT}.

\section{Results with smaller number of samples}
Table~\ref{supp_table4} shows the ablation on amount of the synthetic samples. The performance drops as the number of samples decreases. However, HAST with only 256 samples still performs better than previous methods, such as IntraQ with 45.51\% using 5120 samples.

\begin{table}[ht]
\centering
    \resizebox{0.45\textwidth}{!}{
    \large
    \begin{tabular}{*{10}{c}}
        \toprule
        Amount & IntraQ(5120) & 256 &  1280 & 2560 & 5120 \\
        \midrule
        ACC(\%) & 45.51 & 49.17 & 49.95 & 50.23 & 51.15 \\
        \bottomrule
    \end{tabular}
    }
\caption{Results with smaller number of samples.}
\label{supp_table4}
\end{table}

\end{document}